\documentclass[runningheads]{llncs}

\usepackage[T1]{fontenc}
\usepackage{graphicx}
\usepackage{amsmath}
\usepackage{multirow}
\usepackage{float}
\usepackage{amssymb}
\usepackage{xcolor}

\newcommand{\white}[1]{\textcolor[rgb]{1.0,1.0,1.0}{#1}} 
\newcommand{\tbOK}[1]{\textcolor[rgb]{0.0,0.0,0.0}{#1}}

\begin{document}
\title{\vspace{-65pt}CoralLite: $\mu$CT Reconstruction of\\Coral Colonies from Individual Corallites\vspace{-8pt}}
%
\titlerunning{CoralLite}

%

\author{Jess Jones\inst{1} \and
Leonardo Bertini\inst{1,2,3} \and
Kenneth Johnson\inst{3} \and
\\Erica Hendy\inst{1} \and
Tilo Burghardt\inst{1}}
\authorrunning{J. Jones et al.}
\institute{University of Bristol
\email{\{jess.jones,e.hendy\}@bristol.ac.uk; \email{tilo@cs.bris.ac.uk}} \and
University of Liverpool
\email{leonardo.bertini@liverpool.ac.uk}\and
Natural History Museum
\email{k.johnson@nhm.ac.uk}\vspace{-15pt}}

\maketitle              
\begin{abstract}
\tbOK{The life history of an individual coral is archived within the accreting skeleton of the colony. While reef-forming coral colonies (e.g. massive~\emph{Porites} sp.) may live for hundreds of years and deposit calcareous structures many metres in height and width, their living tissue is a thin outer surface layer comprised of asexually-dividing polyps that only survive a few years. To understand the rate and timing of polyp division and the consequences for colony skeletal growth, scientists need to track the skeletal corallite deposited around each polyp. Here we propose CoralLite, an annotated $\mu$CT scan dataset of entire calcareous skeletons and an associated, first corallite deep learning reconstruction baseline. CoralLite combines fully quantified volumetric segmentations with cross-slice linking for visualisations of 3D models for each corallite up to colony scale. For segmentation, we propose and evaluate in detail a hybrid V-Trans-UNet architecture applicable to segmenting tiled $\mu$CT virtual slabs of~\emph{Porites} sp. colonies. The model is pre-trained on weakly annotated data and topology-aware fine-tuned using fully annotated slice sections with 8k+ manual corallite region annotations. On unseen slices of the same colony, the resulting model reaches 0.94 topological accuracy at mean Dice scores of 0.77 on the same colony and projection axis, and 0.63 mean Dice scores on a different, biologically unrelated specimen. Whilst our experiments are limited in scale and context, our results show for the first time that visual machine learning can effectively support full 3D individual corallite modelling from $\mu$CT scans of coral skeletons alone. For reproducibility and as a baseline for future research we publish our full dataset of 697 $\mu$CT slices, 37 partial or full slice annotations, and all network weights and source code with this paper.\vspace{-5pt}}


\end{abstract}
\vspace{-28pt}
\section{Introduction}
\begin{figure}[t]
\centering\vspace{-8pt}
\includegraphics[width=350px, height=234px]{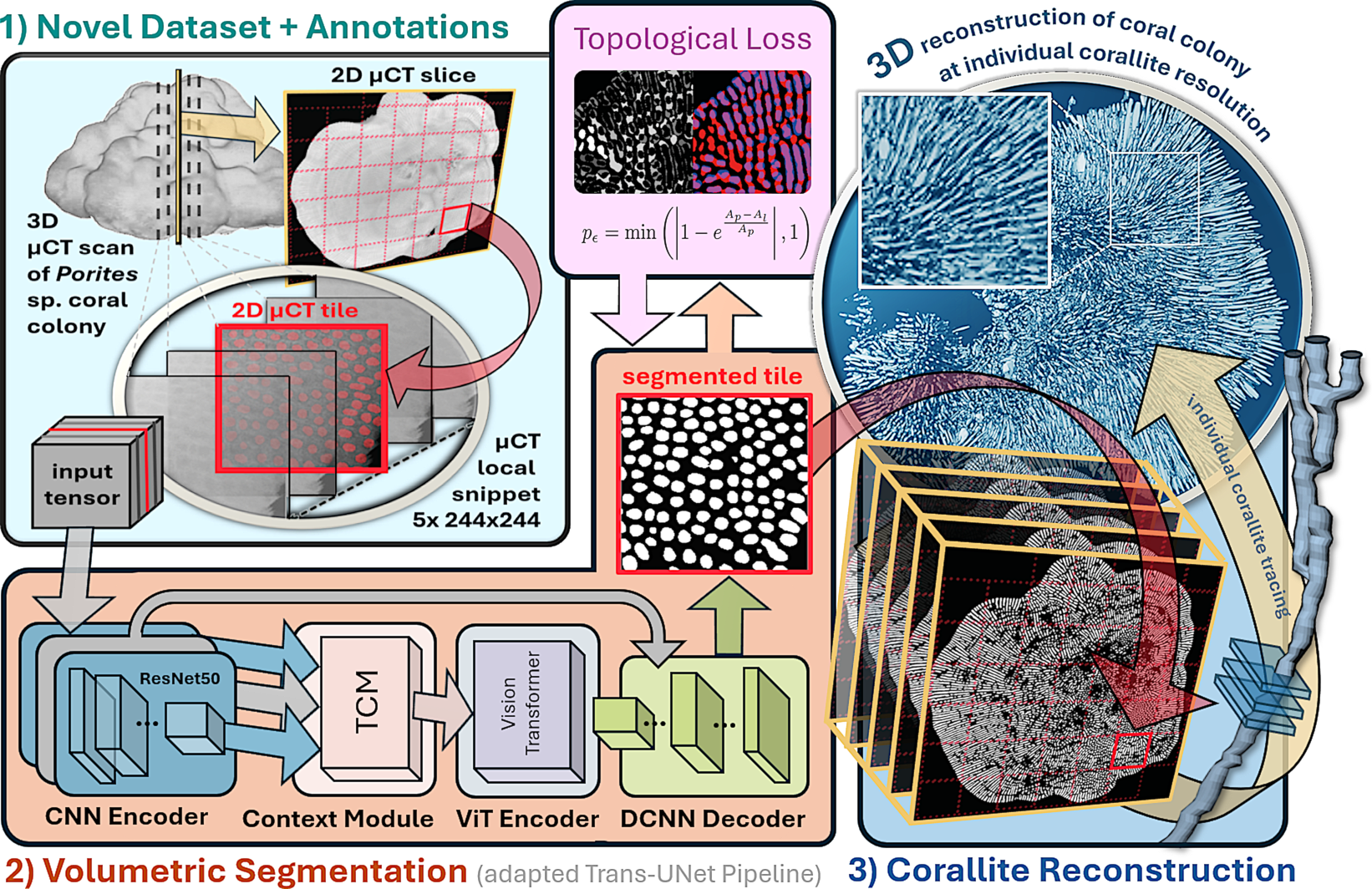}\vspace{-7pt}
\caption{\tbOK{\footnotesize{\textbf{{Overview of CoralLite.}} The proposed pipeline comprises three main stages: \textit{\textbf{(1)}~Novel Dataset:} Sequential 2D cross-section slices from 3D $\mu$CT scans of a \emph{Porites} sp. coral colony are annotated and processed into 5x224x224 tiled snippets. \textit{\textbf{(2)}~Volumetric Segmentation:} A 3D-context-aware Trans-UNet pipeline with a hybrid ResNet50-ViT backbone~(details in Fig.~\ref{model_arch_diagram}) generates corallite segmentations using a topological loss for accurate boundary prediction. \textit{\textbf{(3)}~Corallite Reconstruction:} Segmented tiles are stacked into a volume to generate a 3D corallite-level colony model.}}}\vspace{-10pt}
\label{overview_diagram}
\end{figure}

\vspace{-8pt}
\tbOK{\textbf{Scientific Motivation.} \emph{Scleractinia}, or stony corals, can form colonies by asexual division of individual polyps. Each polyp deposits a surrounding protective tube of aragonite~($CaCO_3$), known as a corallite, which forms a porous wall between neighbouring polyps and collectively creates the lasting skeletal structure of the colony~\cite{darke1993growth,drake2021genetic}.  For example, an individual massive \emph{Porites} sp. colony can grow large enough to form a micro-atoll, or alternatively a dome-shaped skeletal structure that can be over 5 metres in diameter, and encompass over 500~years of growth~\cite{hendy2003chronological,lough1997several}. Fossils of \emph{Scleractinia} date back to the middle Triassic period and have therefore been a feature of ocean ecosystems for more than 230~million years. Since the Miocene, the genus \emph{Porites} has been functionally important in shallow, tropical and sub-tropical waters as one of the key reef-framework builders of warm-water coral reefs~\cite{cabioch2010encyclopedia}.  Massive \emph{Porites} sp. are also the most commonly studied coral for palaeoclimate reconstructions and growth rate variability associated with environmental change~\cite{reefcoreholocene}. In $\mu$CT scans, the structure of these colonies appears as large stacks of cross-sectional slices containing many small interlocked corallites as shown in Figure~\ref{overview_diagram}. Manual tracing of the internal space within individual corallites through these volumes is possible and practised today for fine-grained skeletal analysis, however this approach is time-consuming and laborious~\cite{medellin2022understanding}. \emph{Porites} sp. serve as an excellent case study for analysis automation with direct applicability to vital coral reef research.}

\tbOK{\textbf{Research Aims and Contributions.} This paper considers whether such identity-preserving reconstructions of individual corallites in 3D from $\mu$CT data alone can be automated using deep learning for animal biometrics~\cite{kuhl2013animal}. The approach is outlined in Figure~\ref{overview_diagram}. In contrast to most existing coral reconstruction research, we attempt fully automated segmentation of individual corallites of a scanned colony and quantify baselines of this task here for the first time. Our aim is not to present a complete biological model of coral growth at this stage since expert annotations are limited and expensive to produce at scale. Instead, our aim is to assess whether a reasonably sized, annotated dataset and volumetric learning together are sufficient to support an automated colony reconstruction at corallite level which is useful to practitioners today and sets baselines for the community to build on this work. Recent progress in medical and volumetric image segmentation in particular presented a useful starting point for our work~\cite{cciccek20163d,hatamizadeh2022unetr}. Procedurally, we  trained 3D-aware segmentation models to identify corallite regions encoded in binary segmentation maps, and then apply cross-slice linking procedures supported by topological learning constraints to assemble predicted regions into per-corallite, identity-preserving 3D models. In line with UNet architecture tests for corallites in~\cite{coronel2026leveraging}, we build on hybrid Video-TransUNet architectures and research~\cite{chen2021transunet,zeng2023video} that combines local texture modelling with broader contextual reasoning. Our ResNet--ViT baselines follow this paradigm and operate on stacks of adjacent coral $\mu$CT slices such that cross-slice aggregation can exploit continuity in 3D data. Our contributions are:\vspace{-6pt}
\begin{itemize}
    \item \textbf{Annotated Dataset:} 697~$\mu$CT scan slices plus a set of 8,412 manual corallite ground truth segmentations in full slices of massive~\emph{Porites}  complemented further by point-annotation corallite priors across 33 of the slices;
    \item \textbf{Topology-guided 3D Reconstruction Pipeline:} a volumetric segmentation pipeline based on a V-TransUNet model with cross-slice spatial feature context learning, together with a topological loss for discouraging merged regions with full release of key code and network weights;
    \item \textbf{Baseline Results and Evaluation:} baselines with a first task evaluation across same-colony, cross-axis, and cross-specimen test settings; and 
    \item \textbf{Visualisation Bridge:} a simple cross-slice tracing and geometry generation procedure for reconstructing individual corallites as 3D objects suitable for manual inspection by marine scientists using standard graphics packages;
\end{itemize}}

\vspace{-22pt}
\section{Related Work}\vspace{-9pt}
\label{sec:related}

\tbOK{\textbf{3D Coral Reconstruction in Marine Biology.} Recent research has shifted toward volumetric 3D imaging techniques, such as computed axial tomography (CT) and micro-CT ($\mu$CT), to precisely measure coral extension rates and corallite demography~\cite{yudelman2022coral,medellin2022understanding}. These high-resolution reconstructions allow for a more accurate, non-destructive assessment of individual polyp longevity and skeletal growth trajectories compared to traditional 2D analysis~\cite{benedetti2020polyp,darke1993growth}. By resolving the 3D spatial organisation of coral modules, these methods hold rich, previously inaccessible information about colonial developmental dynamics~\cite{medellin2022understanding}.}

\tbOK{\textbf{Corallite Reconstruction Automation.} Automated computer-vision pipe-lines for individual corallite reconstruction from whole-colony $\mu$CT remain scarce: most coral $\mu$CT work still emphasises manual processing, rather than learned, 3D corallite models. A notable recent exception~\cite{coronel2026leveraging} applies deep semantic segmentation (U-Net family models, including Attention U-Net) to $\mu$CT of scleractinians such as \textit{Montastrea cavernosa} and \textit{Porites astreoides}, separating skeleton from pores to support quantitative 3D analyses without reconstructing corallites. }

\tbOK{\textbf{Cross-Domain Architectural Consideration.} The segmentation of corallites in $\mu$CT data shares several characteristics with medical image segmentation: structures are small relative to the image, neighbouring slices are strongly correlated, and dense annotation is difficult to obtain~\cite{jonsson2023micro,wurfl2016deep}. For such settings, CNNs remain a common baseline, while hybrid CNN--Transformer models have shown that local feature extraction and longer-range context can be combined effectively~\cite{chen2021transunet}. Transformers are attractive in this setting because they can model broader spatial relationships than standard convolution alone, but they are also data-hungry and often benefit from pre-training~\cite{google_vit}. Hybrid architectures help bypass the heavy data requirements of standard Transformers by using a convolutional backbone to handle initial feature extraction. We therefore adopt a ResNet--ViT backbone design for the UNet layout similar to Trans-UNet~\cite{chen2021transunet}.}

\tbOK{\textbf{Temporal Contexts and Topological Guidance.} A second relevant direction is the use of temporal or sequential context modules. Zeng et al.~\cite{zeng2023video} introduce a temporal context module (TCM) for blending information across neighbouring video frames in X-ray segmentation. Although our input is not video, adjacent $\mu$CT slices also contain short-range continuity. We use this idea to encode relationships across slices in a volumetric window. Finally, our downstream task places additional demands on the segmentation output. Standard region-overlap metrics such as Dice are not always sufficient when neighbouring instances must remain separated for corallite reconstruction. This motivates an auxiliary loss designed to penalise merged or over-expanded predicted regions. Similar topologically-motivated constraints have recently proven effective in medical image segmentation; for instance, the architecture in~\cite{shi2023nextou} utilises a binary topological interaction loss to rapidly enforce exclusion and containment relationships among intricate anatomical structures.~\cite{shit2021cldice} proposed clDice, a topology-preserving loss function for tubular structures that is calculated on the intersection of segmentation masks and their morphological skeletons.}

\vspace{-10pt}
\section{Dataset and Setup}\vspace{-7pt}
\label{sec:data}

\begin{figure}[t]
\centering\vspace{-8pt}
\includegraphics[width=346px, height=247px]{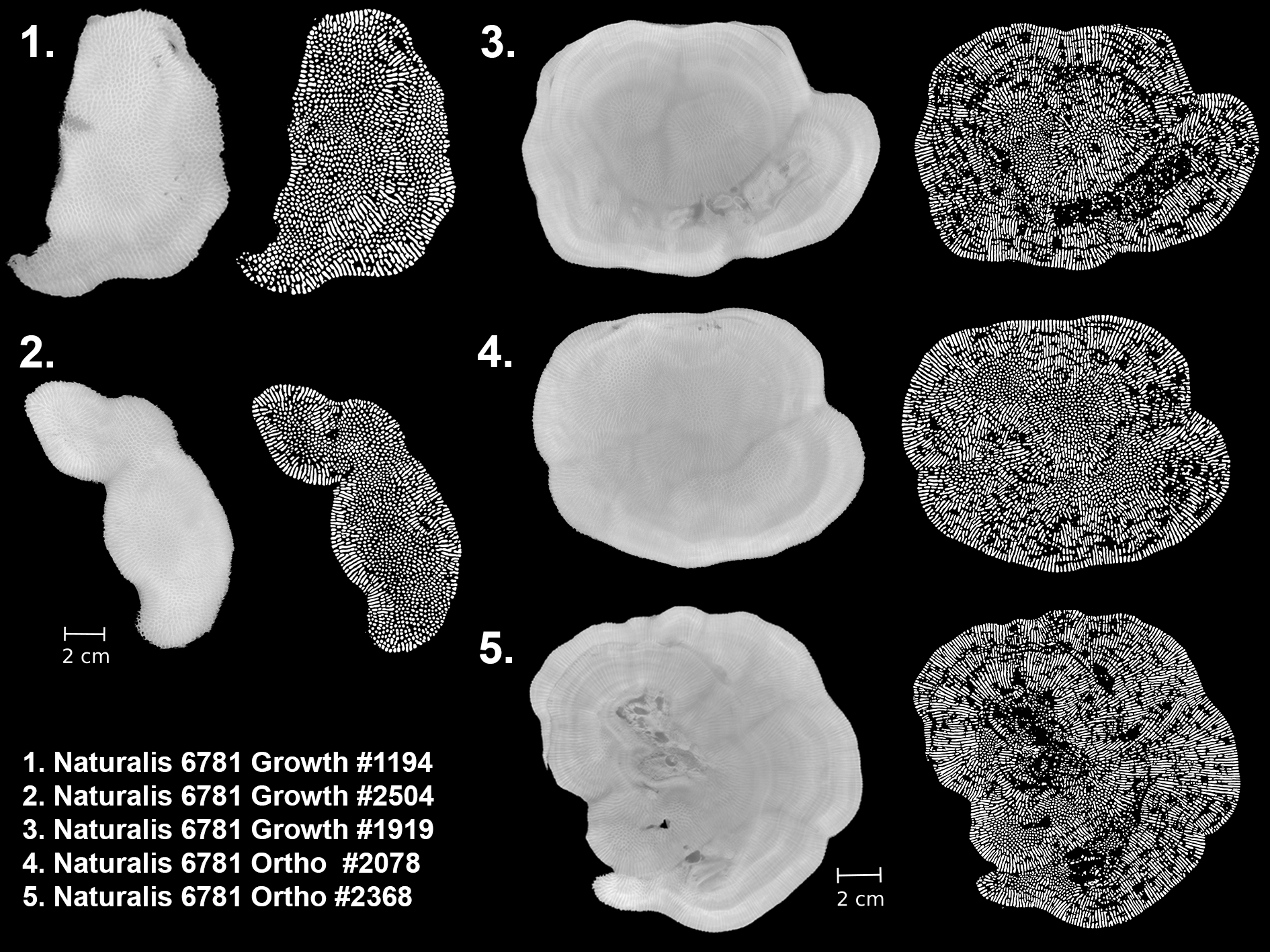}\vspace{-8pt}
\caption{\tbOK{\textbf{{Dataset Sample Slices next to Segmentation Results}}. Shown are 5 of the 697 full-colony slices from the dataset next to segmentation outputs using inference performed on the Naturalis 6781 colony. This specific coral colony was originated from a single coral polyp forming a single corallite.  The coral grew over a five year period into a lumpy hemispheric colony of thousands of corallites. Depicted are samples from both the Growth projection axis, and the Ortho projection axis. These examples illustrate the task's complexity and provide qualitative result visualisations across an entire colony volume and different $\mu$CT projection axes.}}\vspace{-9pt}
\label{inference_ex_image}
\end{figure}

\tbOK{\textbf{Acquisition and Dataset Details.} We use high-resolution $\mu$CT data from three \emph{Porites} specimens from the Natural History Museum (London, UK) and Naturalis Biodiversity Center (Leiden, NL) collections. Specimens were scanned at the Natural History Museum using a Nikon XT H 225 ST system. Reconstructions were exported as 16-bit volumes and further processed into virtual slabs using Avizo 2021.2. The data comprise both weakly annotated~(corallite volume centre points) and fully annotated slices~(full manual binary segmentations of corallite volumes). Weak annotations are produced for a subset of slices from Naturalis 6785 specimen with registration number ZMA.Coel.6785. In addition, we provide four fully annotated slices comprising 8,412 unique corallite interior segments across two \emph{Porites} specimens and two projection settings, that is growth direction and orthogonal direction slices of the volume. The full dataset with all ground truth annotations will be made publicly available with this paper. Figure~\ref{inference_ex_image} visualises 5 of the 697 2D slices of full-colony specimen next to associated corallite segmentation results. Datasets contained in this study are listed in Tab.~\ref{tab:data} and detailed in following four slice sets:\vspace{-4pt}
\begin{itemize}
    \item \textbf{Naturalis 6785}: 60~non-contiguous slices from a growth-axis projection, 33~of which are partially annotated. These data are used for pre-training.
    \item \textbf{Naturalis 6781 Growth}: 248 slices from the growth-axis projection of a different colony (registration number ZMA.Coel.6781). Slice \#1279 is used for fine-tuning and slice \#2499 for evaluation.
    \item \textbf{Naturalis 6781 Ortho}: 388 slices from an orthogonal projection of the same colony. Slice \#1003 is used only for evaluation.
    \item \textbf{Astraeiformis}: single fully annotated evaluation slice of different \emph{Porites} specimen from the Natural History Museum collection (cat.\#1883.11.8.9).\vspace{-5pt}
\end{itemize}}

\label{inference_ex}

\noindent\tbOK{\textbf{Segmentation Task and Evaluation Settings.} The segmentation target is defined straight forwardly as binary corallite volume detection: each pixel is classified as corallite or background. Ground truth dataset slices are represented as 8-bit binary images with 255 and 0 pixel values, respectively. Evaluation is designed to characterise 3 generalisation levels for learned segmentation models:\vspace{-4pt}
\begin{itemize}
    \item \textbf{(Task A) same colony, same axis}: Naturalis 6781 Growth \#2499;
    \item \textbf{(Task B) same colony, different axis}: Naturalis 6781 Ortho \#1003;
    \item \textbf{(Task C) entirely different specimen}: \emph{Astraeiformis}.
\end{itemize}}

\begin{table}
\centering
\caption{\tbOK{\textbf{Summary of Datasets.} Our dataset contribution contains four different slice sets with varying levels of annotation. Partial annotation stipulates corallite centres, whilst full annotation provides exact masks. We use selections from these to perform pre-training, fine-tuning, and different generalisation performance evaluations.}}\vspace{-4pt}
\label{tab:data}
\begin{tabular}{l l c l l}
\hline
\textbf{Specimen} & \textbf{Axis} & \textbf{\white{..}Slices\white{......}} & \textbf{Annotation} & \textbf{Slice Usage} \\
\hline
Naturalis 6785 \white{......}& Growth\white{...} & \white{0}60 & 33 partial & Pre-train \\
Naturalis 6781 & Growth & 248 & Full: \#1279, \#2499\white{.....} & Fine-tune or Test \\
Naturalis 6781 & Ortho & 388 & Full: \#1003 & Test \\
Astraeiformis & -- & \white{00}1 & Full slice & Test \\
\hline
\end{tabular}
\end{table}

\vspace{-14pt}
\section{Methods and Experiments}\vspace{-6pt}
\label{sec:method}

\subsection{Volumetric Input Generation}
\begin{figure}[h]
\centering
\includegraphics[scale = 0.8]{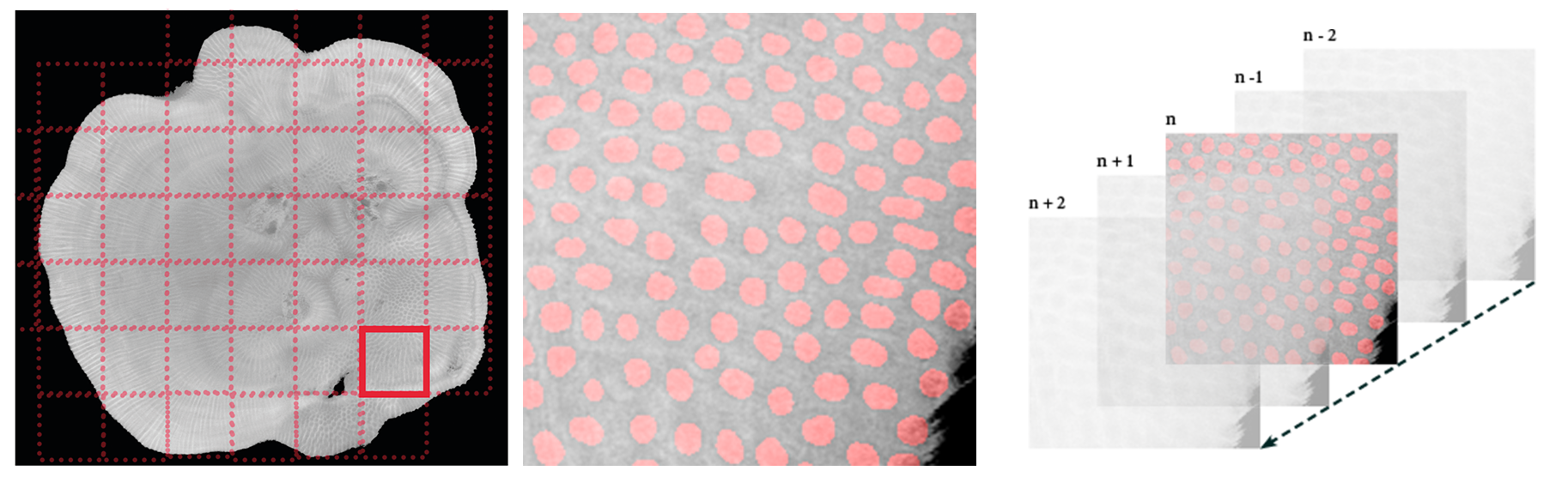}\vspace{-8pt}
\caption{\tbOK{\textbf{Volumetric Input Tensor Generation.} Tiled volumetric stacking provides local context from adjacent slices for segmentation. Left to right: Illustrative example of raw image tiling with window $224 \times 224$ and step size $224$; An example composite tile with annotation overlay in red, from the Naturalis 6785 dataset; An example snippet of dimensions $5 \times 224 \times 224$. Note that slice $n$ is situated in the centre of the volume.}}
\label{inputs}
\end{figure}

\vspace{-2pt}
\label{subsec:inputs}
\tbOK{\textbf{Sliding-Window Tile Extraction.} Each $\mu$CT slice has high spatial resolution, while individual corallites occupy only a small number of pixels. To avoid  downsampling and fit network tensor dimensions, we process each slice using tiled crops of size $224 \times 224$~(see Figure~\ref{inputs}). Tiles are generated by a sliding window with step size $k$. During pre-training on Naturalis 6785 we use $k=224$, while during fine-tuning on Naturalis 6781 Growth \#1279 we use $k=50$ for overlap.}

\tbOK{\textbf{Cross-Slice Volumetric Stacking.} To exploit continuity across adjacent slices, each training sample is constructed as a snippet of neighbouring tiles: $\mathbf{x} \in \mathcal{R}^{D \times H \times W}$,
where $H=W=224$, $D$ is odd, and the annotation corresponds to the centre slice.
We adopt a $D=5$ regime following the methodology of \cite{zeng2023video} to capture local volumetric context. This depth provides a balance between capturing the longitudinal trajectory of corallites and minimising the memory footprint during training. Figure~\ref{inputs} illustrates the tiling process and resulting volumetric inputs.}

\subsection{Segmentation Pipeline}\vspace{-2pt}
\label{subsec:model}

\begin{figure}[t]
\centering\vspace{-19pt}
\includegraphics[scale = 0.3]{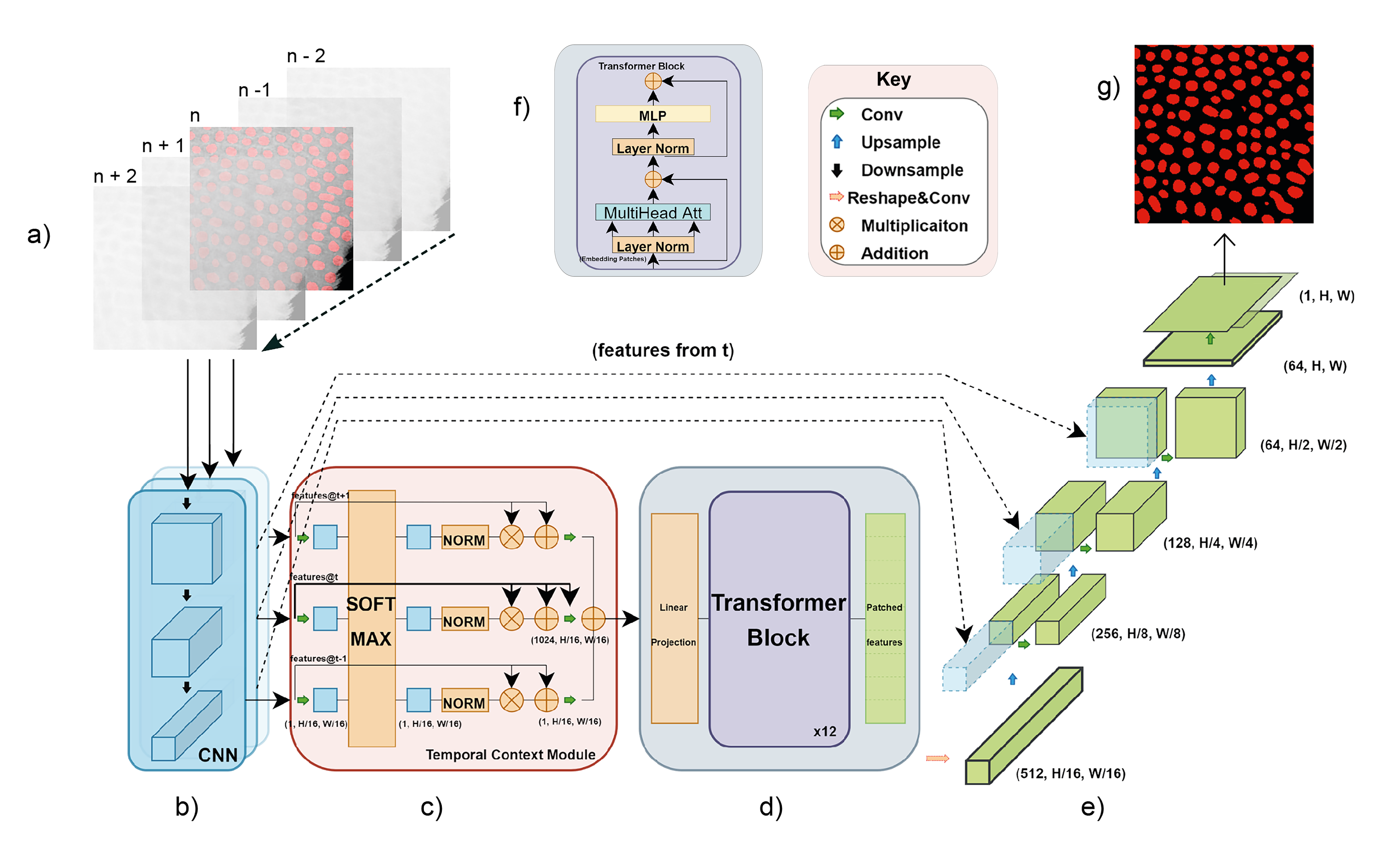}\vspace{-10pt}
\caption{\footnotesize{\tbOK{\textbf{Full Architecture Diagram}. Our pipeline uses a Video-TransUNet~\cite{zeng2023video} design but applied here to \emph{spatial} volume analysis. It utilises CNN feature encoding with short-range volumetric learning and complex spatial context modelling via a ViT before DCNN decoding: \textbf{\textit{(a)}}: $5 \times 224 \times 224$ input tensors are provided to \textbf{\textit{(b)}}: a pre-trained ResNet-50 CNN encoder with $3$x skip connections. \textbf{\textit{(c)}}: Feature encodings are blended across the volumetric dimension before being passed to \textbf{\textit{(d)}}: a pre-trained 12~layer ViT with Multi-head Attention and \textbf{\textit{(f)}}: MLP transformer block components. \textbf{\textit{(e)}}:~Finally, through cascaded DCNN up-sampling incorporating skip information a final segmentation output of size $224 \times 224$ is produced in \textbf{\textit{(g)}}. (adapted Video Trans-UNet of~\cite{zeng2023video})}}}\vspace{-10pt}
\label{model_arch_diagram}
\end{figure}

\tbOK{\textbf{Hybrid Trans-UNet Architecture.} We adapt the Video-TransUNet design of~\cite{zeng2023video} to snippets formed as tiled $\mu$CT slabs. We symmetrise and reinterpret Video-TransUNet's temporal dimension as spatial depth. Our model then  combines a ResNet-50 encoder, a temporal context module (TCM)~\cite{zeng2023video}, a 12-layer ViT, and a cascaded decoder with skip connections as shown in Figure~\ref{model_arch_diagram}.}

\tbOK{\textbf{Stage-by-stage Feature Processing.} Given an input stack $\mathbf{x} \in \mathcal{R}^{D \times H \times W}$, the ResNet encoder extracts per-slice features. The attention-based context module then considers features across the depth dimension, allowing neighbouring slices to contribute local volumetric context. The resulting representation is passed to the ViT encoder, which models longer-range spatial dependencies before decoding using a deconvolutional DCNN to yield a segmentation mask output $f(\mathbf{x}) \in \mathcal{R}^{H \times W}$ for the processed tile, i.e. the centre slice of~$\mathbf{x}$.}

\vspace{-11pt}
\subsection{Topology-aware Loss Regime}\vspace{-4pt}
\label{subsec:topoloss}
\tbOK{\textbf{Motivation for Connectivity Penalty.} For downstream single corallite reconstruction, it is important that neighbouring corallites remain separated in the predicted masks. Standard overlap losses such as binary cross-entropy and Dice do not directly penalise some important failure modes, such as thinly merged neighbouring corallite volumes or over-expanded false positive regions.}

\tbOK{\textbf{Component Penalty Formulation.} Using basic connected-component analysis on binary annotation and prediction maps via \texttt{skimage.measure}, we obtain the set of predicted regions~$P$ and labelled regions~$L$. For each predicted connected component~$p \in P$ with area~$A_p$, we identify its nearest (potentially overlapping) labelled region~$l \in L$ with area~$A_l$ and define the component penalty:\vspace{-2pt}
\begin{equation}
p_{\epsilon} = \min\left(\left|1 - e^{\frac{A_p - A_l}{A_p}}\right|, 1\right).\vspace{-12pt}
\end{equation}}

\begin{figure}[t]
\centering\vspace{-6pt}
\includegraphics[scale = 0.55]{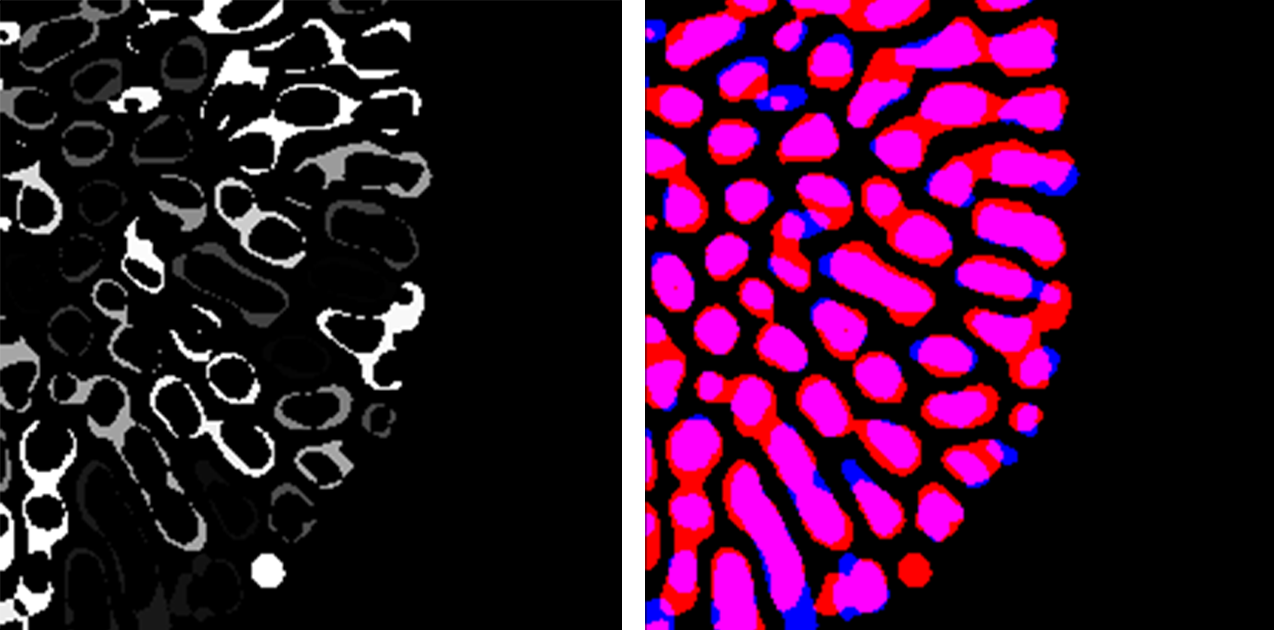}
\includegraphics[scale = 0.233]{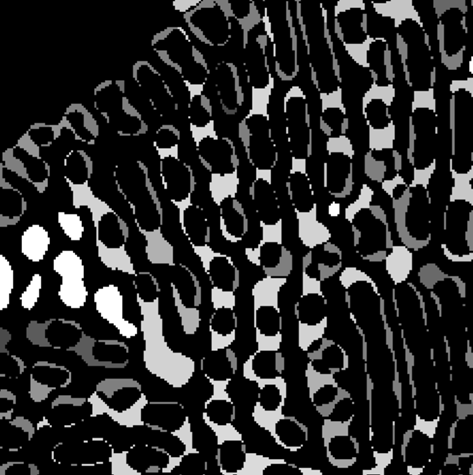}
\includegraphics[scale = 0.233]{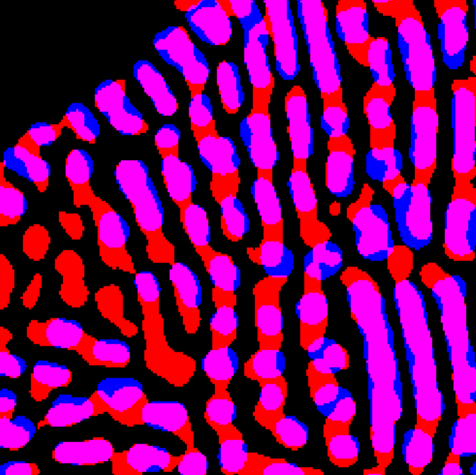}
\vspace{-8pt}
\caption{\tbOK{\textbf{Topological Error Maps.} Two example depictions of connectivity-driven error maps used for training loss calculation highlighting incorrectly connected or generated corallite regions. The topological error map~$\mathcal{E}$ is presented on the left of each image pair, while composite ground-truth (blue) and prediction (red) maps are given on the right of each pair. The grayscale error map demonstrates how the topological loss can specifically target both false positives (isolated red pixel groups) and merged regions~(bridged by pixels shown in red), encouraging structural separation specifically, which is vital for accurate downstream 3D tracing of individual corallites.}}\vspace{-7pt}
\label{topo_loss_vis2}
\end{figure}

\tbOK{\textbf{Error Map and Topological Loss Stipulation.} Each pixel belonging to a region~$p$ is assigned the corresponding penalty if it does not belong to the matched ground-truth region~$l$. This yields an error map~$\mathcal{E}(n)$ that highlights for each pixel~$n$ its contribution towards incorrect corallite connections, false positive corallites and over-expanded regions~(see Figure~\ref{topo_loss_vis2}). This map is used to calculate a topological loss by averaging evidence over all~$N$ pixels of a tile during training as:\vspace{-8pt}
\begin{equation}
\mathcal{L}_{\text{Topo}} =
\frac{1}{2N}\sum_{n=1}^{N}
\sigma\left(\mathcal{E}(n)\right)-0.5. 
\end{equation}}

\vspace{-12pt}
\subsection{Definition of Learning Objectives}
\label{subsec:trainingobj} \vspace{-4pt}

\tbOK{\indent\textbf{Pre-Training.} Since weakly annotated pre-training data~(see Tab.~\ref{tab:data}) lacks sufficient fine-grained information for connectivity analysis we utilise standard Binary Cross-Entropy~\cite{BCE} and Dice~\cite{dice} losses without topological awareness as bulk pre-training objective:
\vspace{-10pt}
\begin{equation}
\mathcal{L} = \frac{1}{2}\left(\mathcal{L}_{\text{BCE}} + \mathcal{L}_{\text{Dice}}\right).
\vspace{-8pt}
\end{equation}}

\tbOK{\textbf{Training Loss Mixture.} During main fine-tuned training of the network---where manually annotated pixel-specific ground truth is available---we add the topological loss with $T=0.1$ after a warm-up period of 50 epochs:
\vspace{-1pt}
\begin{equation}
\mathcal{L} = \frac{1}{2}\left(\mathcal{L}_{\text{BCE}} + \mathcal{L}_{\text{Dice}}\right) + T\mathcal{L}_{\text{Topo}}.
\end{equation}}
\vspace{-25pt}

\begin{figure}[t]
\centering
\vspace{-7pt}
\includegraphics[width=160px]{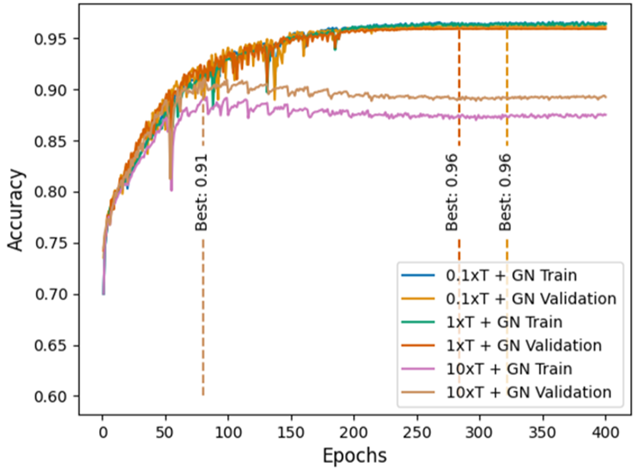}\white{.}
\includegraphics[width=180px]{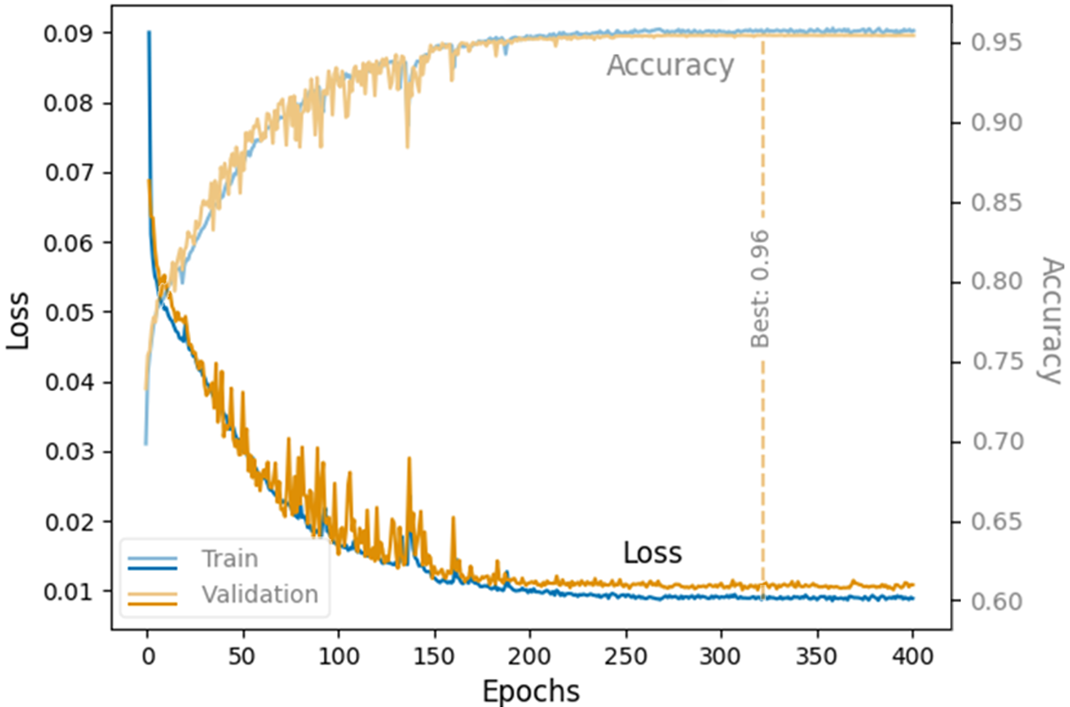}
\vspace{-19pt}
\caption{\tbOK{\textbf{Topological Contribution Tuning, Training and Validation Plots}. Performance development during main fine-tuning of our model. \textbf{\textit{(left)}}~A comparison of different topology coefficients~$T$ confirms stable performance around $0.1$-$1.0$. \textbf{\textit{(right)}}~With peak validation accuracy of~0.96 and saturated loss reduction at 320 epochs at a stable plateau, the depicted model (FT VLrg 0.1xT + GN in Tab.~\ref{tab:result}) has learned to segment unseen validation corallite slice data without significant over-fitting.}}\vspace{-12pt}
\label{tv_graph}
\end{figure}

\label{sec:experiments}
\subsection{Network Training}\vspace{-4pt}

\tbOK{\textbf{Two-Stage Optimisation with Augmentation.} Learning is happening in two stages. First, the segmentation model is pre-trained on the weakly annotated Naturalis 6785 data where ResNet-50 + ViT are initialised with standard ImageNet weights. Secondly, the pipeline is fine-tuned in a main training run on a fully annotated slice from the target colony, Naturalis 6781 Growth \#1279. All experiments are conducted with Adam, learning rate $5\times10^{-4}$, and batch size 5. In both stages, 10\% of the available training data are held out for validation. The remaining training tiles are augmented by random rotation, horizontal and vertical flips, and, in later experiments, additive Gaussian noise. We use a snippet depth of 5, fix context module parameters after pre-training, and set the topological loss coefficient to $T=0.1$. Figure~\ref{tv_graph} depicts the full training curves.}

\vspace{-9pt}
\section{Segmentation Results and Evaluation}
\label{sec:results}
\vspace{-5pt}
\subsection{Summary of Performance Baselines}\vspace{-4pt}
\tbOK{\textbf{Segmentation Metrics for Quantification.} Segmentation performance is evaluated using the mean Dice Similarity Coefficient~(mDSC)~\cite{dice} as standard primary segmentation metric alongside a task-specific Topology Score~(TS) defined as $1-\mathcal{L}_{\text{Topo}}$ for approximating reproduction success of connected structures essential for downstream corallite tracing.}

\tbOK{\textbf{Segmentation Performance.} Table~\ref{tab:result} summarises the main quantitative corallite segmentation results across all settings. Fine-tuning substantially improves all over the pre-trained baseline across all metrics. Performance differences based on hyperparameter choices are fully documented, but are relatively small, showing the stability and suitability of the adapted V-Trans-UNet for the task at hand. The overall best-performing configuration in our experiments combines dense tiling~($k=50$) during fine-tuning, topology-aware loss at~($T=0.1$), and additional augmentation with Gaussian noise. On the target tasks, this model reaches mDSC scores of 0.77 on \textbf{(Task A)} Naturalis 6781, 0.71 on \textbf{(Task B)} Naturalis 6781, and still cross-species performance 0.63 on \textbf{(Task C)} \emph{Astraeiformis}.}

\begin{table}[t]\vspace{-4pt}
\label{res_table} \caption{\tbOK{\textbf{Corallite Segmentation Performance Baselines}. Full testing performance baselines for all models across the three generalisation tasks (see Sec.~\ref{sec:data}) using \textbf{mDSC}: Mean Dice similarity, and \textbf{Topo}: Topological accuracy: $1 - \mathcal{L}_{Topo}$. The full model with maximum tile overlap, topological loss utilisation and Gaussian Noise augmentation generalises best and offers best, or joint best, performance across all testing domains. Best or shared best performance with the full model is highlighted in bold.}}\vspace{-4pt}
    \centering
    \scriptsize
   \begin{tabular}{|l|l||c|c||c|c||c|c|}
\hline
\multicolumn{1}{|c|}{\multirow{4}{*}{\normalsize{\textbf{Model}}}} & & \multicolumn{2}{c||} {\white{\large{\textbf{|}}}\small{\textbf{Task A}}} & \multicolumn{2}{c||}{\white{\large{\textbf{|}}}\small{\textbf{Task B}}} & \multicolumn{2}{c|}{\white{\large{\textbf{|}}}\small{\textbf{Task C}}} \\
& & \multicolumn{2}{c||}{\white{..}\textbf{(same axis)}\white{.}} & \multicolumn{2}{c||}{\white{.}\textbf{(same species)}\white{.}} & \multicolumn{2}{c|}{\white{.}\textbf{(diff. species)}\white{.}} \\
& & \multicolumn{2}{c||}{\white{.}Naturalis 6781\white{.}} & \multicolumn{2}{c||}{\white{.}Naturalis 6781\white{.}} & \multicolumn{2}{c|}{\white{.}\emph{Astraeiformis}\white{.}} \\
 & & \multicolumn{2}{c||}{Growth \#2499} & \multicolumn{2}{c||}{Ortho \#1003} & \multicolumn{2}{c|}{} \\
\cline{3-8}
 \white{\large{\textbf{|}}} & \white{-}Parameters\white{-}&\white{.}mDSC\white{.} & Topo & \white{.}mDSC\white{.} & Topo & \white{.}mDSC\white{.} & Topo \\
\hline \hline 
\textbf{Pre-training only}\white{...}\white{\normalsize{|}} & \white{-}n/a & 0.008 & 0.66 & 0.01 & 0.75 & 0.01 & 0.68 \\
\hline
\textbf{FT} (fine-tuned)\white{\normalsize{|}} & \white{-}$k=224$ & 0.75 & 0.93 & 0.67 & 0.92 & 0.61 & 0.88 \\ 
 & \white{-}$k=112$ & 0.76 & 0.94 & 0.69 & 0.93 & 0.57 & 0.80 \\ & \white{-}$k=50$  & 0.76 & 0.94 & 0.71 & 0.93 & 0.56 & 0.79 \\
\hline
\textbf{FT} \textbf{+ $\mathcal{L}_{Topo}$}\white{\normalsize{|}} & \white{-}$T=0.01$ & 0.76 & 0.93 & 0.65 & 0.92 & 0.59 & 0.85 \\ 
\textbf{no tile overlap} (k\white{.}=\white{.}224) & \white{-}$T=0.1$ & 0.76 & 0.93 & 0.64 & 0.92 & 0.59 & 0.85 \\ 
 & \white{-}$T=1$ & 0.76 & 0.93 & 0.62 & 0.90 & 0.60 & 0.85 \\ 
 & \white{-}$T=10$  & 0.76 & 0.93 & 0.65 & 0.92 & 0.59 & 0.85 \\ 
 & \white{-}$T=100$  & 0.77 & 0.93 & 0.62 & 0.90 & 0.60 & 0.85 \\ 
\hline
\textbf{FT} \textbf{+ $\mathcal{L}_{Topo}$}\white{\normalsize{|}} & \white{-}$T=0.01$\white{-} & 0.77 & 0.94 & 0.70 & 0.93 & 0.51 & 0.80 \\ 
 \textbf{max tile overlap} (k\white{.}=\white{.}50)& \white{-}$T=0.1$  & 0.77 & 0.94 & 0.70 & 0.93 & 0.52 & 0.80 \\ 
& \white{-}$T=1$  & 0.77 & 0.94 & 0.70 & 0.93 & 0.51 & 0.79 \\ 
& \white{-}$T=10$ & 0.76 & 0.94 & 0.69 & 0.93 & 0.48 & 0.80 \\ 
& \white{-}$T=100$  & 0.76 & 0.94 & 0.69 & 0.93 & 0.53 & 0.81 \\ 
\hline
\textbf{FT} \textbf{+ $\mathcal{L}_{Topo}$}\white{\normalsize{|}} &  \white{-}$T=0.01$ & 0.76 & 0.94 & 0.70 & 0.92 & 0.61 & 0.87 \\
 \textbf{max tile overlap} (k\white{.}=\white{.}50)& \white{-}$T=0.1$ & \textbf{0.77} & \textbf{0.94} & \textbf{0.71} & \textbf{0.93} & \textbf{0.63} & 0.87 \\
 \textbf{plus noise augmentation}\white{....}& \white{-}$T=1$ & 0.77 & 0.94 & 0.70 & 0.93 & 0.60 & 0.86 \\
& \white{-}$T=10$ & 0.77 & 0.94 & 0.70 & 0.93 & 0.63 & \textbf{0.89} \\
\hline
\end{tabular}\vspace{-12pt}
\label{tab:result}
\end{table}

\subsection{Detailed Segmentation Evaluation}\vspace{-8pt}
\tbOK{\textbf{Fine-Tuning is Critical.} Without fine-tuning, the pre-trained model performs poorly on all evaluation slices. Expectedly, only detailed training annotations that reflect the exact corallite shape are sufficient to learn how to segment them given the visual complexity of coral colony scans. However, fine-tuning on one fully annotated slice is sufficient to recover strong performance on unseen, same colony data, and transfers reasonably to a different projection axis and specimen.}

\begin{figure}
\centering
\includegraphics[scale = 0.75]{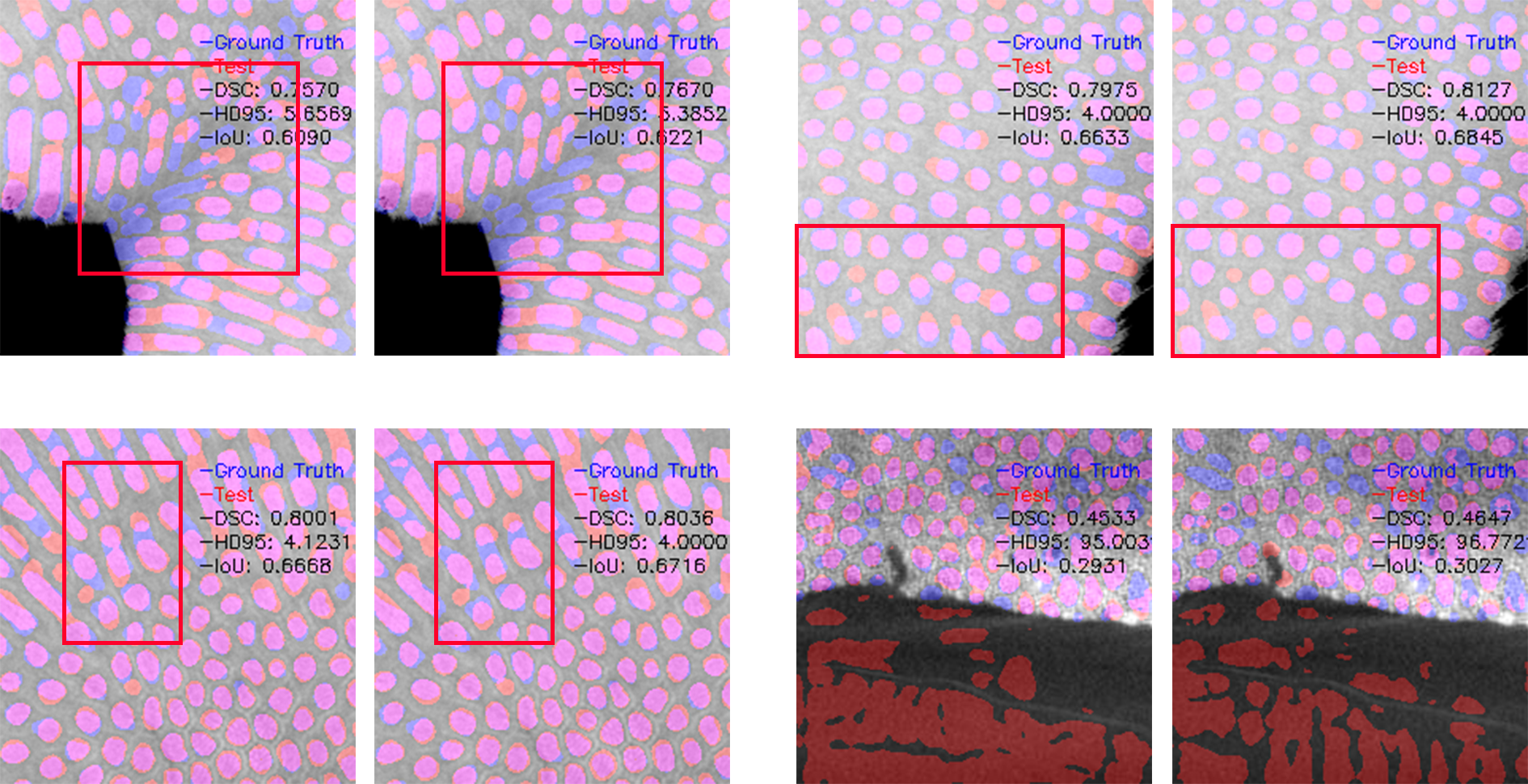}\vspace{-8pt}
\caption{\tbOK{\textbf{Connectivity Impact of Topological Loss.} Four composite example pairs of predictions (red) and ground-truth data (blue) superimposed on coral tiles, that is (left) without and (right) with topological loss utilised ($k=50, T=1$). Note the improvement in structural corallite separation, particularly in areas highlighted by the red bounding box. \textbf{Bottom right pair}: Note that scanning noise and xray echoes, here on a detail of the \emph{Astraeiformis} slice, lead to major out-of-domain~(OOD) misinterpretation and false positives independent of topological loss considerations. Detection of OOD content is one key direction of future work in this domain~and wider AI research.}}\vspace{-30pt}
\label{tloss_preds_ex}
\end{figure}

\tbOK{\textbf{Qualitative Effect of Topology-aware Loss \& Augmentation.}
Adding the topological loss does not produce large changes in mDSC, but improves the separation and shape of neighbouring predicted regions. Whilst the footprint in the overall Topo quantification shown in Tab.~\ref{tab:result} is minimal due to the averaging nature of the overall loss shown, the practical effect is visible in qualitative comparisons~(e.g. see Figure~\ref{tloss_preds_ex}), where connected false positives between adjacent corallites are reduced. For instance, when testing for Task B at $k=50$, utilising $\mathcal{L}_{Topo}$ reduces the absolute corallite count error by~$2\%$ (measured with $(T=0.1)$). The effect is most consistent for densely tiled fine-tuning settings at values of $T\leq 1$).  Performance on the \emph{Astraeiformis} slice testing cross-species performance is lower than on the Naturalis 6781 slices throughout. Adding Gaussian noise augmentation alone improves \emph{Astraeiformis} mDSC by up to $8\%$~(compare final two row blocks of Tab.~\ref{tab:result}), while preserving performance on same-colony test slices. Qualitative inspection suggests that false positives in regions with scanner noise or echos are particularly improved.}

\subsection{3D Corallite Reconstruction}\vspace{-4pt}
\label{subsec:reconstruction}

\tbOK{\textbf{Basic Cross-Slice Tracing.} In order to turn 2D corallite segmentation into 3D corallite tube reconstructions suitable for visualisation of entire colonies and for manual inspection, for each connected region $R_i$ in slice $S_n$ we search for a matching counterpart region $R_j$ in the previous slice $S_{n-1}$. A match is accepted based on centroid distance and intersection-over-union~(IoU) satisfying 
$d(R_i, R_j) < \gamma$ and $\mathrm{IoU} (R_i, R_j) > \beta$. Thresholds can be altered for task-specific visualisations -- visuals in this paper are generated with~$\gamma = 0.3$ and $\beta = 0.3$. If a match is found, $R_i$ is assigned to the same corallite object as $R_j$; otherwise, a new object trace is started.}

\begin{figure}[t]
\centering\vspace{-11pt}
\includegraphics[scale = 0.81]{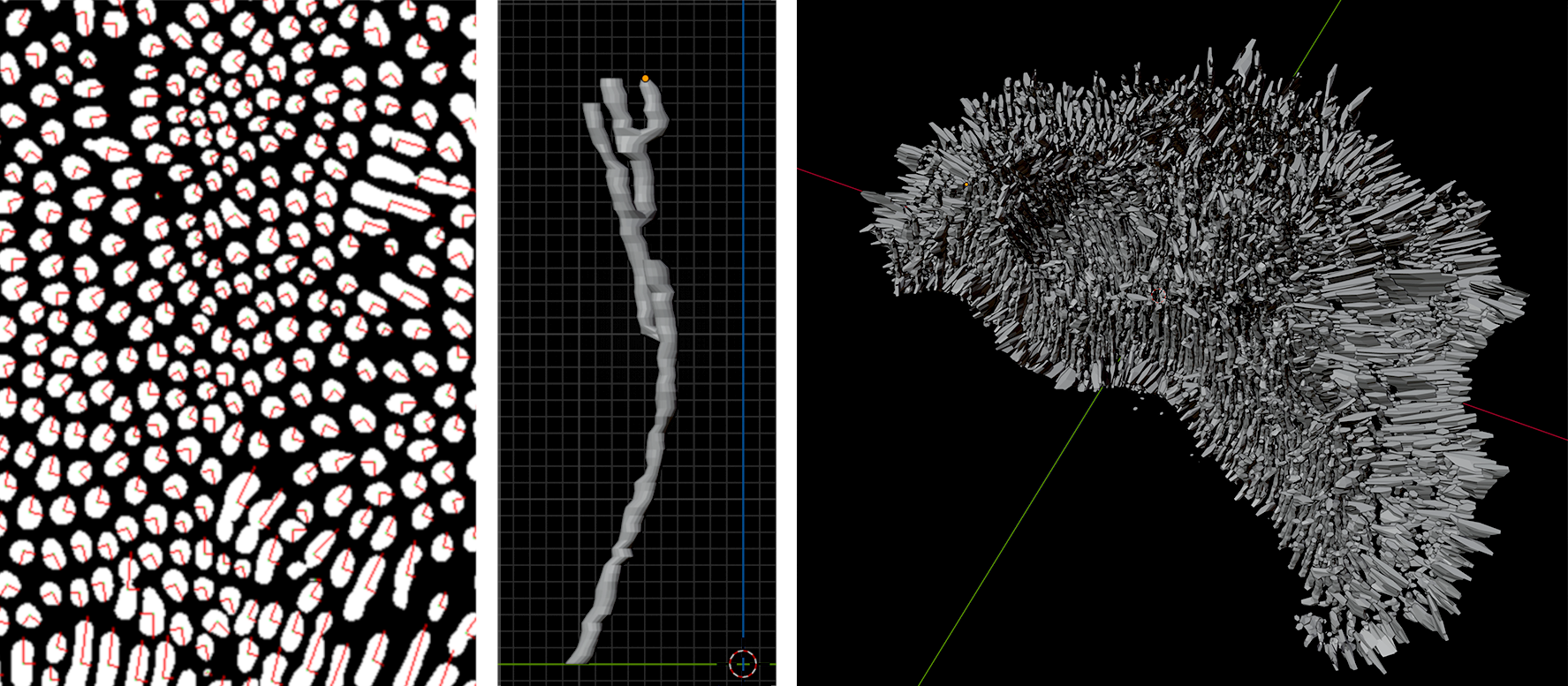}\vspace{-9pt}
\caption{\tbOK{\textbf{Corallite Reconstruction and 3D Visualisation.} \textbf{\textit{(left)}} Per-slice corallite representation based on centroid, major and minor axis lengths, and in-plane
orientation shown as vector pair. \textbf{\textit{(middle)}}~Corallite \#87 as viewed along the X-axis where centroid tracing has translated stacks of 2D segments into a coherent 3D structure for visualisation that captures the natural geometry, curvature, split and termination as well as scale of the corallite. \textbf{\textit{(right)}}~Model of reconstructed colony viewed along the Z-axis via corallites \#0 - \#20,000. Note how this first individual corallite visualisation at colony scale highlights key coral features such as directional alignment of local corallite groups, dense packing of tubes, and the overall outward growth of the colony.}}\vspace{-14pt}
\label{reconstruction}
\end{figure}

\tbOK{\textbf{3D Colony Visualisation based on Corallite Objects.} Each traced region is represented by its centroid, major and minor axis lengths, and in-plane orientation~(see Figure~\ref{reconstruction}, left). These parameters define an ellipse in 3D space at a depth corresponding to slice index. Ellipses belonging to the same track are then connected along the slice axis to form corallite object models, which can then be visually inspected as individual corallites~(see Figure~\ref{reconstruction}, middle) or in large groups up to the context of a full colony volume~(see Figure~\ref{reconstruction}, right). The resulting reconstructions capture corallite tube structures, alignment and growth directions within the colony and many other features. Whilst such visualisations at colony scale have never been attempted before, we emphasise that this is for inspection only. A full scientific 3D quantification against manually traced ground truths remains a subject for future work, as discussed in Section~\ref{sec:discussion}.}

\section{Brief Discussion}\vspace{-8pt}
\label{sec:discussion}
\tbOK{\textbf{First Proof-of-Concept Pipeline.} The results suggest that volumetric segmentation via deep learning architectures can indeed provide a workable basis for automatic, individual-level corallite reconstruction from $\mu$CT data alone. In particular, pre-training on weak annotations and fine-tuning on just one fully annotated slice from the target colony were sufficient to obtain useful segmentation quality on unseen slices from the same colony. The resulting masks were adequate for generating interpretable 3D corallite object models at colony scale performed and visualised here for the first time.}

\tbOK{\textbf{Assumptions, Constraints and Future Work.} Since our fully annotated dataset of manually labelled 8,412 corallite regions is relatively limited in size, reported results only provide performance baselines, but imply by no means performance caps on the tasks investigated. Furthermore, the full 3D reconstruction is not scientifically quantified here and is generated for first interpretable visualisations only. Further, our cross-specimen results suggest that out-of-domain application remains a practical challenge. Thus, this study provides first datasets and baselines to enable and inspire research on full scientific coral colony reconstructions -- it does not yet claim to provide an imageomics application that can reliably estimate corallite demography indicators such as budding rate variability and polyp lifespan as a function of colony age. This paper lays down the first foundational work for this via public datasets, annotations, and comparison baselines for the community. That is, further research is required to completely automatically produce fully accurate and scientifically quantified 3D corallite-level reconstruction of coral colonies from CT scans alone.} \vspace{-10pt}

\begin{figure}[t]
\centering\vspace{-7pt}
\includegraphics[scale = 0.56]{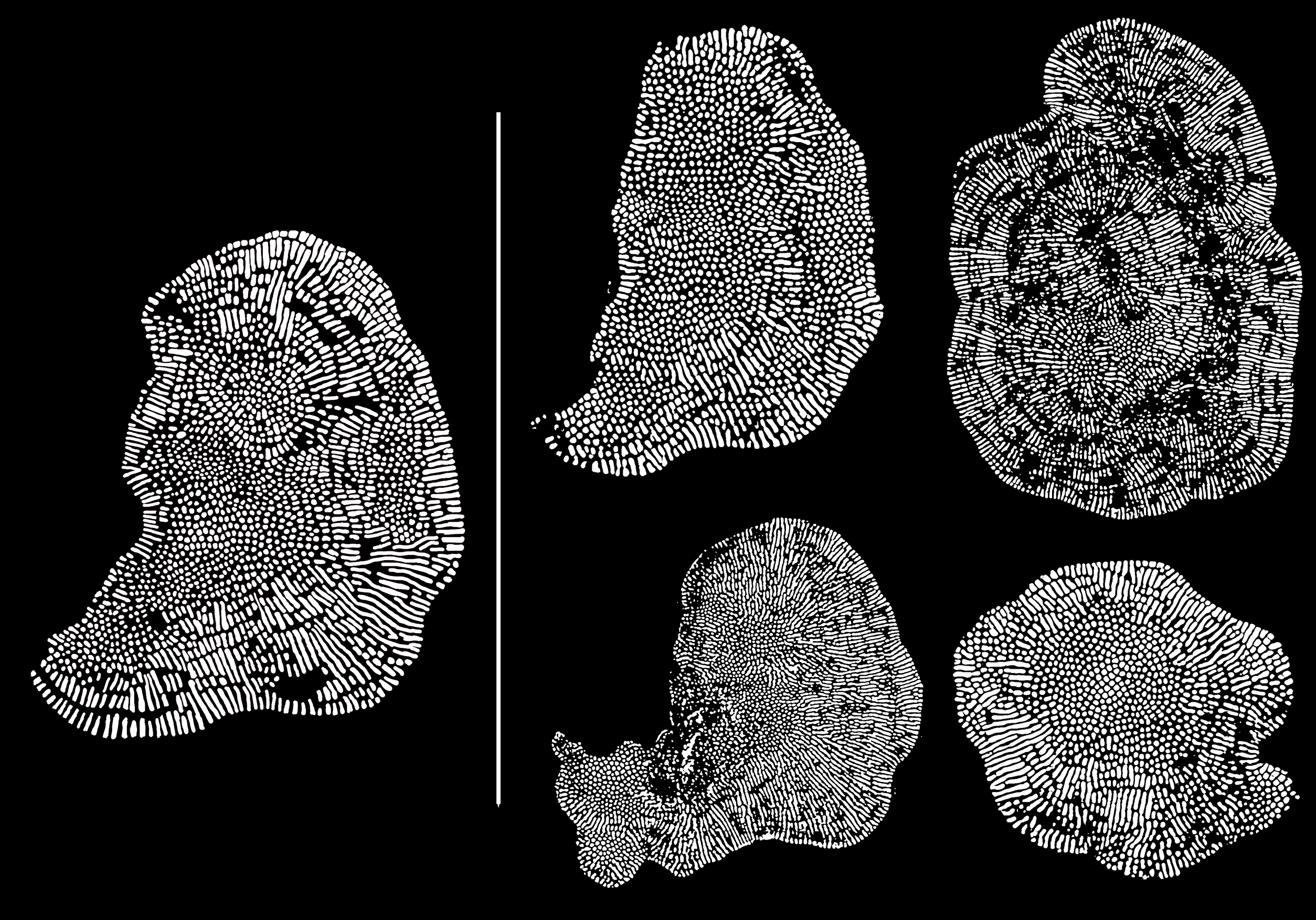}\vspace{-8pt}
\caption{\textbf{Further Segmentation Results.} Examples from Naturalis 6781 colony; \textbf{\textit{(left)}} Annotation example, slice 1279 (Growth). \textbf{\textit{(right)}} Inference outputs, clockwise from top-right: slice 2043 (Ortho), 2718 (Ortho), 1379 (Growth), and 0119 (Growth).}\vspace{-10pt}
\label{seg}
\end{figure}

\begin{figure}[t]
\centering
\includegraphics[scale = 0.365]{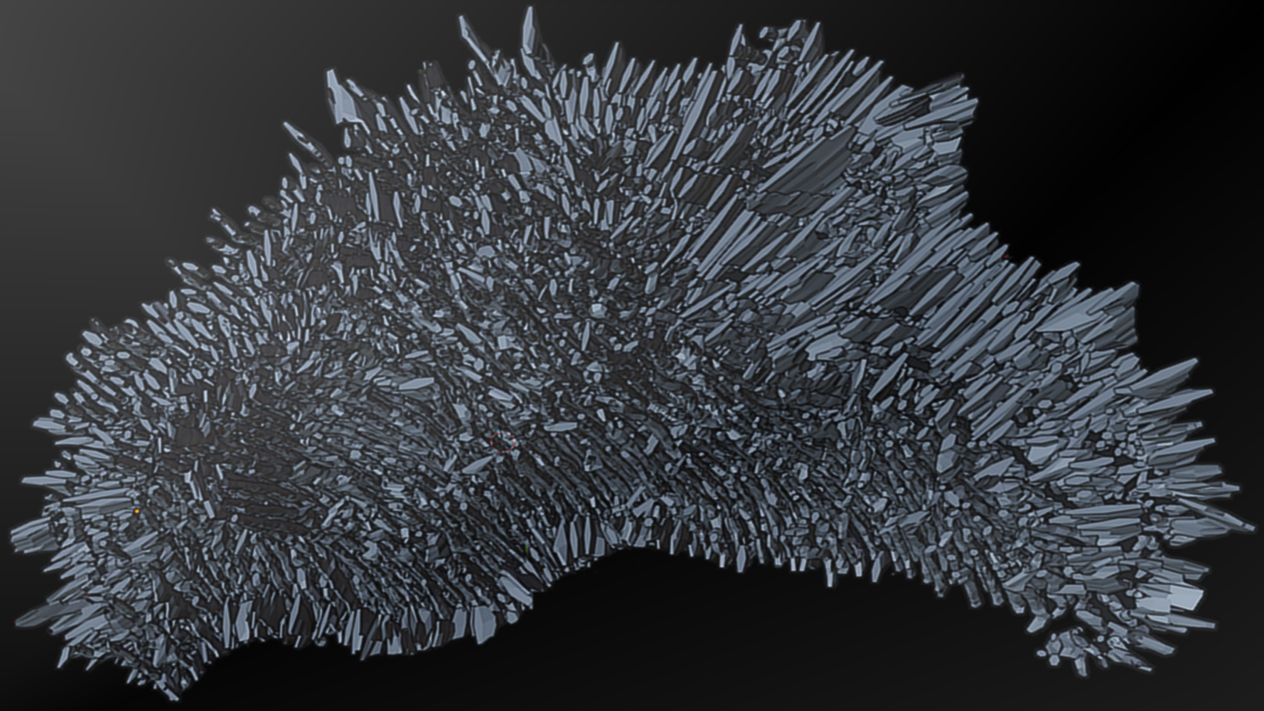}\vspace{-8pt}
\caption{\textbf{Example of Colony-scale Reconstruction (Close-up Visualisation).} }\vspace{-12pt}
\label{re}
\end{figure}

\section{Conclusion}\vspace{-10pt}
\tbOK{We presented CoralLite, a new $\mu$CT coral dataset, with individual corallite annotations and a V-Trans-UNet inspired deep learning pipeline for corallite segmentation and reconstruction, together with full segmentation baselines. We produced a first approximate colony-scale coral visualisation made from generated individual corallite objects. The underlying segmentation model was fully evaluated on manually annotated test data. Our hybrid V-Trans-UNet inspired architecture combined short-range cross-slice feature blending with a topological loss aimed at discouraging merged neighbouring regions. In our experiments, fine-tuning on one fully annotated slice from a target colony was shown sufficient to produce useful segmentation quality on unseen data. The resulting masks supported an automatic 3D corallite reconstruction process at colony scale. The presented system establishes a proof-of-concept that fully automated reconstructions of corallites are feasible at colony level. It provides useful visualisations of colonies and marks a first step towards imageomic automation for understanding the rate and timing of polyp division and the consequences for colony skeletal growth. We publish the full CoralLite dataset of 697 $\mu$CT slices,  37 partial or full slice annotations with 8,412 manual corralite region labellings, and all network weights and pipeline source code with this paper. We hope that this can enable and inspire the computer vision for conservation community to engage in furthering research in this area in order to understand corals better and find new ways to safeguard the remaining reefs of the planet.}

\section{Acknowledgements}
J.J. and T.B. would like to thank and acknowledge the support of the University of Bristol HPC facilities.  E.H. and L.B. gratefully acknowledge the dedicated labelling of corallites on scan slices completed by an enthusiastic team of University of Bristol Environmental Geoscience undergraduates; Arisha Fancy, Poppy Hammond, Sharifah Nor Qistina Syed Izuan, Louisa Lou, Emma Mody, and Khalil Imran Abdul Rahman, who were funded through the NERC GW4 DTP Undergraduate Research Experience Placement scheme and the University of Bristol School of Earth Sciences Summer Internship programme.

Support for L.B. and this study is also acknowledged through 4D-REEF, a Marie Skłodowska-Curie Innovative Training Network funded by European Union Horizon 2020 research and innovation programme under grant agreement number 813360. We are also grateful for additional support from SYNTHESIS+ Transnational Access project NL-TAF-TA3-009, the Natural History Museum Science Investment Fund and the University of Bristol Covid Recovery Fund.

\bibliographystyle{splncs04}
\bibliography{references.bib}
\end{document}